\documentclass[11pt]{article}
\usepackage{listings}
\usepackage{hyperref}
\usepackage[ruled,vlined,linesnumbered,algo2e]{algorithm2e}
\usepackage{subcaption}
\usepackage{caption}
\usepackage{enumitem}
\usepackage{microtype}
\usepackage{graphicx}
\usepackage{xcolor}
\usepackage{booktabs} 
\usepackage{multirow}
\usepackage{amsmath}
\usepackage{amssymb}
\usepackage{mathtools}
\usepackage{amsthm}
\theoremstyle{plain}
\newtheorem{theorem}{Theorem}[section]

\theoremstyle{definition}

\newtheorem{assumption}[theorem]{Assumption}
\theoremstyle{remark}
\newtheorem{remark}[theorem]{Remark}

\usepackage[utf8]{inputenc} 
\usepackage[T1]{fontenc}    
\usepackage{hyperref}       
\usepackage{url}            
\usepackage{booktabs}       
\usepackage{amsfonts}       
\usepackage{nicefrac}       
\usepackage{microtype}      
\usepackage{xcolor}         
\usepackage{wrapfig}
\usepackage[preprint]{acl}
\usepackage{times}
\usepackage{latexsym}

\usepackage[T1]{fontenc}

\usepackage[utf8]{inputenc}

\usepackage{microtype}
\usepackage{inconsolata}

\usepackage{graphicx}
\usepackage{lipsum}

\title{OSGR: Observation-Support Graph Reasoning for Earth Observation Question Answering}

\author{
  Dazhou Yu\thanks{Equal contribution.} \\
  Emory University
  \And
  Zerong Wang\footnotemark[1] \\
  Georgia State University
  \And
  Riyang Bao \\
  Emory University
  \AND
  Longxiang Li \\
  Emory University
  \And
  Yang Liu \\
  Emory University
  \And
  Liang Zhao\thanks{Corresponding author.} \\
  Emory University
}

\begin{document}
\maketitle
\begin{abstract}

Answering geographic queries over Earth observation (EO) archives requires solving a joint retrieval-and-derivation problem under strict spatiotemporal constraints. Existing LLM agents fail to ground observation items or suffer from search-space explosion. We propose OSGR, a two-stage framework that factorizes this search into offline symbolic concept deduction and online Relation-Conditioned Grounding. OSGR leverages an offline Observation-Support Graph to deduce valid target-to-result support paths at the concept level, and applies spatiotemporal compatibility kernels to instantiate grounded observation-item sets. On Earth-Bench and DisasterM3-Reason, OSGR improves GPT-4o accuracy from 49.3\% to 63.7\% while reducing the execution-step ratio from 2.9 to 2.5.
\end{abstract}

\section{Introduction}
Earth observation (EO) archives---which store massive amounts of raster data (observation items) such as satellite imagery, climate variables, and reanalysis products---are essential for understanding our planet \citep{mai2024opportunities}. Answering questions using these archives supports consequential applications including disaster response, climate and environmental monitoring, food security, water management, and urban planning \citep{feng2025earthagent}. However, this task is difficult because there is a wide gap between natural-language information needs and the low-level observation items stored in the archive \citep{janowicz2019sosa,ogc2023oms}.

Recently, geospatial foundation models and Large Language Models (LLMs) have been used for generic geographic question answering \citep{mai2021geographic,zhang2024bb}, but vanilla prompting relies entirely on parametric knowledge and lacks a verifiable grounded observation-item set. To address this, retrieval-augmented generation (RAG) and Graph RAG methods inject external text documents, metadata, or relational context into the reasoning loop \citep{fan2024survey,liang2025geographrag}. However, these methods treat EO observation items as plain text or static graph nodes. They ignore the physical structures of observation items, meaning they cannot find a verifiable, accurate set of observation items to answer the query. To interact with observation items directly, tool-augmented agentic systems have recently emerged to interleave multi-step reasoning with active environment execution and dedicated EO toolchains \citep{feng2025earthagent}. Unfortunately, without a clear structural goal to limit their search, general agents must look through an explosive number of possible observation items and operations. This often leads to critical execution failures, such as selecting the wrong observation items, misaligned time windows, wrong locations, or redundant tool sequences \citep{bao2026spatialagent}.

\begin{figure}[t]
\centering
\includegraphics[width=0.9\linewidth]{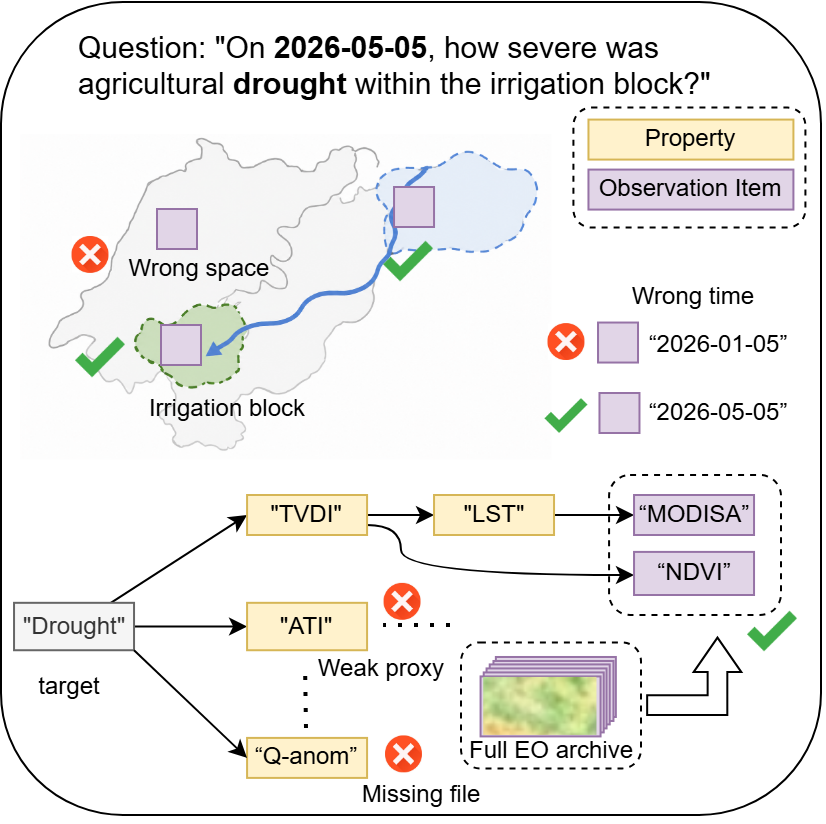}
\caption{A drought-stress EO question.}
\label{fig:intro-drought}
\end{figure}

The fundamental breakdown of existing systems stems from three intrinsic, compounding challenges that connect human logic, observation-item search, and physical compatibility:
\textbf{(1) Scientific Target-to-Result Derivation:} Human queries use abstract environmental phenomena, but archives only store raw or derived physical measurements. Connecting them requires strict scientific derivation, formalized as a support path (e.g., figuring out agricultural drought from a dryness index calculated from Land Surface Temperature (LST) and NDVI, as shown in Fig.~\ref{fig:intro-drought}). LLMs often fail to generate these valid support paths from memory.
\textbf{(2) Massive Search Space of Observation Items:} Even if a valid support path is identified, the physical archive contains too many candidate observation items from different sensors, resolutions, and times, as shown in Fig.~\ref{fig:intro-drought}. Searching through the unorganized observation items creates a huge combinatorial search problem for agents.
\textbf{(3) Dynamic Spatiotemporal Compatibility:} In the physical world, spatial and temporal validity depends heavily on the specific relationship required by the query. Time lags, spatial coverage, and connectivity (like river flows) impose different strict constraints \citep{allen1983maintaining,randell1992spatial,freeman2007hydrologic}. As shown in Fig.~\ref{fig:intro-drought}, ignoring these constraints leads to physical mismatches that invalidate the entire reasoning.

To address these challenges, we present \textbf{Observation-Support Graph Reasoning (OSGR)}, a framework that reformulates EO question answering as a task of \textit{graph-constrained grounded observation-item set selection} using a two-stage factorization. Instead of searching blindly, OSGR separates logical scientific derivation from observation-item verification.

The core of our framework is an offline \textbf{Observation-Support Graph}, hereafter referred to as the support graph, built on the ISO/OGC OMS standard \citep{ogc2023oms}. 
This graph organizes messy physical observations into a clear three-layer structure: Target, Property, and Result nodes. 
Given a specific query, OSGR dynamically extracts target-to-result \textit{support paths}. This stage answers "what should be derived" before looking at any observation items. It restricts reasoning to documented scientific relations and prunes the downstream candidate space before observation-item grounding, addressing Challenge 1 and Challenge 2. 
However, a support path is just an abstract blueprint until it matches real-world observation items. 
In the second stage, OSGR performs \textbf{Relation-Conditioned Grounding} to ground these support paths to observation items. 
Using spatial and temporal compatibility kernels, OSGR filters out observation items that do not match the required locations, times, or resolutions. 
This stage answers "which specific observation items can satisfy the support path" for the query, solving Challenge 3. 
By factorizing the task and conditioning the final answer on both the logical reasoning chain and verified observation items, OSGR reduces spatiotemporal mismatches and improves grounded answering.

In summary, our core contributions are fourfold:
\begin{itemize}[leftmargin=*,wide]
\setlength\itemsep{-0.2em}
    \item We present an EO-specific task formulation that casts question answering as \textit{graph-constrained grounded observation-item set selection}, systematically coupling observation items, scientific support paths, and spatiotemporal topology. More broadly, this treats retrieval as selecting a relation-constrained executable evidence set rather than independently relevant passages.
    \item We construct an ISO/OGC OMS-aligned \textit{Observation-Support Graph} as a cognitive scaffold to organize heterogeneous concepts and dynamically extract query-conditioned scientific support paths, pruning the downstream combinatorial candidate space.
    \item We propose a \textit{Relation-Conditioned Grounding} mechanism equipped with spatial and temporal compatibility kernels to ground support paths to candidate observation items in the archive graph, enforcing strict spatiotemporal compatibility.
    \item Extensive evaluations on Earth-Bench and DisasterM3-Reason demonstrate that OSGR consistently outperforms state-of-the-art baselines in answer accuracy, support path correctness, and spatiotemporal feasibility.
\end{itemize}
\section{Related Work}
\paragraph{EO and geospatial foundation models.}
Foundation models have increasingly been adapted to geospatial and EO settings, where language understanding must interact with places, spatial concepts, and gridded observations. GeoLLM extracts geospatial knowledge from LLMs, while BB-GeoGPT adapts language models with GIS-specific corpora and instructions \citep{manvigeollm,zhang2024bb}. Broader GeoAI studies further emphasize the need to handle spatial semantics, heterogeneous data, and domain-specific reasoning \citep{mai2024opportunities}. In EO, Earth-Agent and related MLLM-based systems expand from single-image perception toward EO products, spectral data, and tool interfaces \citep{feng2025earthagent}. These works strengthen geospatial knowledge, visual understanding, and tool access, but they do not explicitly model observation items from observable result concepts to answer-supporting concepts.

\paragraph{Retrieval-augmented and graph-based geospatial reasoning.}
Retrieval-augmented generation injects external evidence into LLMs, and graph-based RAG extends this idea by retrieving relational context such as entities, paths, or subgraphs \citep{lewis2020retrieval,fan2024survey,edge2024local}. In geospatial settings, Spatial-RAG combines spatial filtering with semantic retrieval, while GeoGraphRAG retrieves geospatial modeling subgraphs to guide workflow and code generation \citep{yu2025spatialrag,liang2025geographrag}. Such methods improve relevance and interpretability, but EO observation selection is not only retrieval over text, POIs, or geoprocessing steps. EO archives are heterogeneous across products, footprints, time intervals, variables, and preprocessing requirements; a retrieved subgraph is not necessarily an observation-support path; and retrieved nodes still need grounding to concrete observation items with valid spatiotemporal support. Unlike standard GraphRAG, OSGR retrieves a scientific support path and then instantiates it as an executable observation-item set whose members jointly satisfy relation-specific physical constraints.

\paragraph{Agentic geospatial and EO analysis.}
Tool-using LLM agents such as ReAct and Reflexion interleave reasoning with external actions, motivating geospatial systems that collect data, generate workflows, or execute EO analysis \citep{yao2023react,shinn2023reflexion}. GeoGPT, GeoTool-GPT, and LLM-Find improve geospatial tool use and data retrieval, while Earth-Agent provides a structured EO tool ecosystem and Spatial-Agent grounds geo-analytical questions in GIScience concepts and executable workflow graphs \citep{zhang2024geogpt,wei2024geotoolgpt,ning2025autonomous,feng2025earthagent,bao2026spatialagent}. These systems show the importance of tool access and structured intermediate representations, but general agents still search over large action spaces without an explicit grounded-observation objective. OSGR instead constructs support paths using an OMS-aligned graph \citep{ogc2023oms,janowicz2019sosa} and grounds them to concrete observation items under relation-conditioned spatial and temporal constraints.

\section{Problem Formulation}
We formulate Earth observation question answering as a retrieval-and-derivation problem over large scale archives. 
Let $V_{\mathrm{arc}}=\{x_1,x_2,\dots\}$ denote the global Earth observation archive, each \textit{observation item} (an EO coverage unit) $x\in V_{\mathrm{arc}}$ is formalized as a tuple:
\[
x=(\Omega_x,T_x,\nu_x,M_x),
\]
where $\Omega_x$ is the spatial footprint, $T_x$ is the acquisition epoch, $\nu_x$ is the physical variable, and $M_x$ is the metadata (e.g., resolution). 
These items induce an \textit{archive graph} $G_{\mathrm{arc}}=(V_{\mathrm{arc}},E_{\mathrm{arc}})$.
Edges in $E_{\mathrm{arc}}$ link observation items through spatial or temporal proximity relations such as overlap or lag.
For a natural-language query $q$, we write $q\mapsto(a_q,s_q,\tau_q)$, where $a_q$ is the abstract target environmental phenomenon of interest (e.g., "drought"), $s_q$ is the spatial cue that denotes the geographic boundary, and $\tau_q$ is the temporal cue. 
Answering the query $q$ requires identifying a feasible family of observation-item sets, denoted as $\mathcal{X}^{\theta}_q$, where each set $X \in \mathcal{X}^{\theta}_q$ ($X \subseteq V_{\mathrm{arc}}$) corresponds to one variable type. The task is formulated as finding the family of subsets satisfying:
\begin{equation}
\label{eq:problem}
\begin{aligned}
\mathcal{X}^{\theta}_q&=\{X\subseteq V_{\mathrm{arc}}\mid  S(X,q)\ge\theta_q\},\\
S(X,q)&=F(X,q)\bigl[C_s(X,s_q)+ C_t(X,\tau_q)\bigr],
\end{aligned}
\end{equation}
where $S(X,q)$ is the compatibility score, $\theta_q$ is a query-dependent threshold. $F(X,q)$ evaluates whether $X$ contains usable variables for the required derivation and checks execution-level compatibility (e.g., sensor resolution),  $C_s$ and $C_t$ score spatial and temporal compatibility. A final answer is generated from $\mathcal{X}^{\theta}_q$ together with the support paths connecting the selected variables to $a_q$.

Existing methods cannot solve this problem effectively due to the following challenges:
(1) {Scientific derivation:} the query target $a_q$ is an abstract phenomenon, the system must verify a valid derivation from $a_q$ to the variables contained in $\mathcal{X}^{\theta}_q$. 
(2) {Combinatorial search:} constructing $\mathcal{X}^{\theta}_q$ from the archive graph $G_{\mathrm{arc}}$ creates an explosive search space over products, locations, and times. 
(3) {Dynamic spatiotemporal compatibility:} the scores $C_s(X,s_q)$ and $C_t(X,\tau_q)$ must enforce query- and derivation-dependent spatial and temporal compatibility, rather than static metadata matching.

\section{OSGR for Earth Observation Question Answering}
\begin{figure*}[t]
\centering
\includegraphics[width=\textwidth]{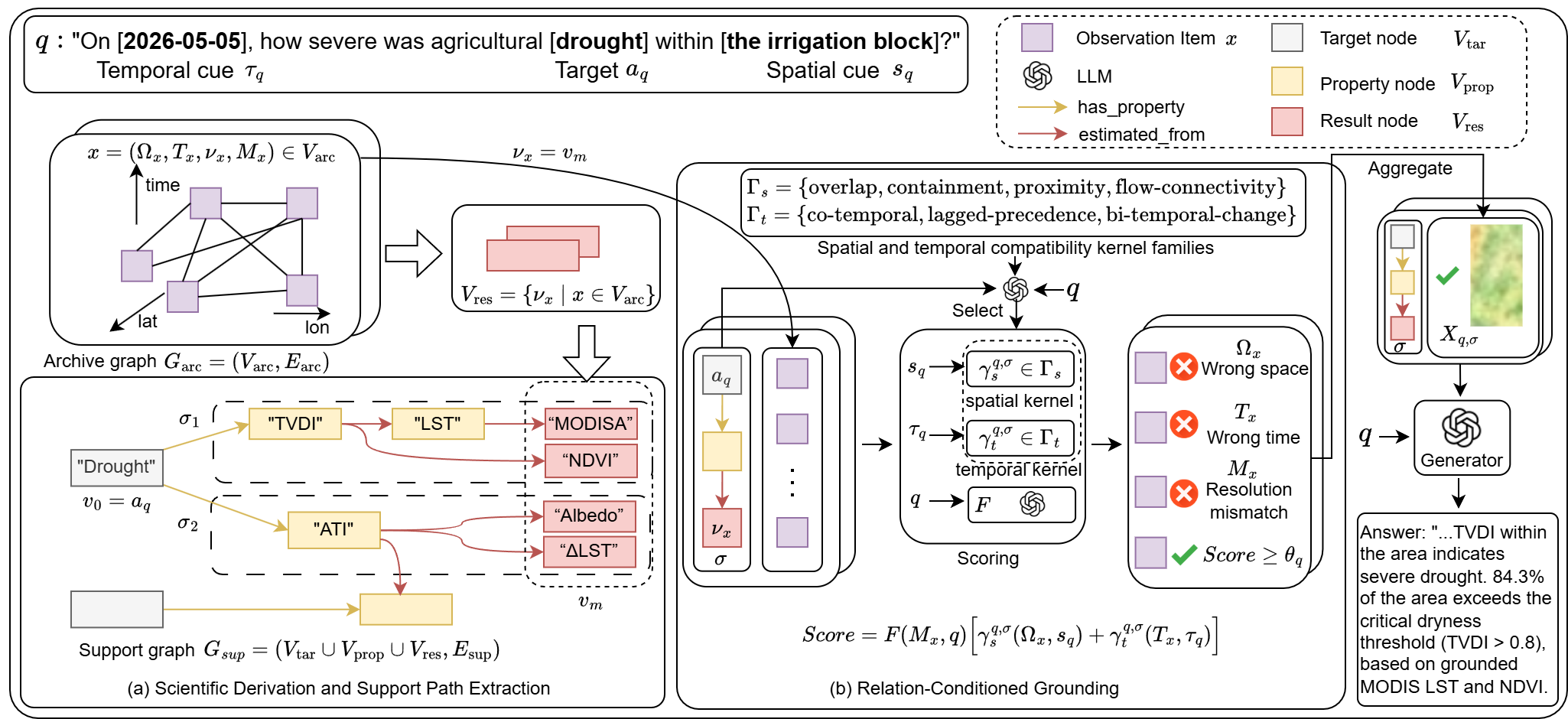}
\caption{Illustration of OSGR.  Given a query $q$ containing target, spatial, and temporal cues, we (a) dynamically extract target-to-result support paths from support graph $G_{\mathrm{sup}}$. (b) Each support path is grounded to candidate observation items in the archive graph $G_{\mathrm{arc}}$ using spatiotemporal compatibility kernels and metadata verification.}
\label{fig:osgr-arch}
\end{figure*}
\subsection{Overview}
Answering complex Earth observation (EO) queries requires bridging the steep semantic gap between abstract human phenomena (e.g., "agricultural drought") and concrete, low-level gridded rasters. 
As illustrated in our framework overview (Fig.~\ref{fig:osgr-arch}), OSGR factorizes this challenge into a strategic two-stage pipeline: a \textit{knowledge-driven concept stage} (as detailed in Section~\ref{sec:kg-construction}) and a \textit{physics-driven spatiotemporal grounding stage} (as detailed in Section~\ref{sec:path-selection}). 

First, OSGR leverages the pre-constructed, offline support graph representing rigorous scientific derivations over domain concepts. Given a specific user query, we dynamically extract a set of query-conditioned target-to-result support paths from this static graph (Fig.~\ref{fig:osgr-arch}(a), Section~\ref{sec:kg-construction}). This step decouples qualitative environmental phenomena from concrete observation items, effectively narrowing down the conceptual search space.
Second, we ground these support paths to candidate observation items in the archive graph $G_{\mathrm{arc}}$ (Fig.~\ref{fig:osgr-arch}(b), Section~\ref{sec:path-selection}). 
Using relation-conditioned spatial and temporal compatibility kernels coupled with metadata verification, OSGR filters out physical mismatches to identify a precise, feasible grounded observation-item set. 
This grounded observation-item set is finally aggregated and fed into a natural language generator to synthesize a highly accurate, quantified answer.

\subsection{Scientific Derivation and Support Path Extraction}
\label{sec:kg-construction}
To address the scale sensitivity of observation-item retrieval, OSGR decouples abstract scientific reasoning from direct archive search. 
Specifically, we utilize the offline, domain-specific support graph to model physical priors at the conceptual level. 
Following the ISO/OGC Observations, Measurements, and Samples (OMS) conceptual framework \citep{ogc2023oms}, we model this domain prior as a tripartite directed graph that reflects physical measurement hierarchies: Target (the studied environmental phenomenon), Property (the mediating physical quantities), and Result (the digital measurement types). 
The support graph is constructed offline from EO scientific documentation; details are provided in Appendix~\ref{app:support-graph-construction}.
As shown in Fig.~\ref{fig:osgr-arch}(a), we formalize the support graph $G_{\mathrm{sup}}$ as:
\begin{equation}
\label{eq:support-graph}
\begin{aligned}
G_{\mathrm{sup}} &=(V_{\mathrm{sup}},E_{\mathrm{sup}}),\\
V_{\mathrm{sup}} &=V_{\mathrm{tar}}\cup V_{\mathrm{prop}}\cup V_{\mathrm{res}},
\end{aligned}
\end{equation}
where $V_{\mathrm{res}} = \{\nu_x\mid x\in V_{\mathrm{arc}}\}$ denotes normalized, archive-available Earth Observation (EO) variable types (e.g., result node NDVI in Fig.~\ref{fig:osgr-arch}(a)) rather than volatile individual observation items. 
The set of valid relational transitions $E_{\mathrm{sup}}$ is defined directly over these domains:
\begin{equation}
\label{eq:support-edges}
\begin{aligned}
E_{\mathrm{sup}} \subseteq &\ \bigl(V_{\mathrm{tar}}\times\{\texttt{has\_property}\}\times V_{\mathrm{prop}}\bigr) \\
&\cup \bigl(V_{\mathrm{prop}}\times\{\texttt{estimated\_from}\}\times V_{\mathrm{prop}}\bigr) \\
&\cup \bigl(V_{\mathrm{prop}}\times\{\texttt{estimated\_from}\}\times V_{\mathrm{res}}\bigr).
\end{aligned}
\end{equation}
By matching structural walks to the inverse mathematical flow of remote-sensing retrievals, this schema restricts derivation to documented, typed scientific relations before observation-item ingestion.

However, the global support graph $G_{\mathrm{sup}}$ remains static and contains pathways irrelevant to a specific user query $q$. 
To prune the conceptual search space, we extract a query-conditioned projection $\mathcal{S}_q$ containing target-to-result support paths that link the query target $a_q \in V_{\mathrm{tar}}$ to archive-available result variable types:
\begin{equation}
\label{eq:support-paths}
\begin{aligned}
\mathcal{S}_q
&=\{\sigma=(v_0,\ldots,v_m)\mid v_0=a_q,\ v_m\in V_{\mathrm{res}},\\
&\quad \forall \ell=1,\ldots,m,\
\exists r:\ (v_{\ell-1},r,v_\ell)\in E_{\mathrm{sup}}\}.
\end{aligned}
\end{equation}
Under this formulation, each extracted $\sigma$ is a minimal, schema-consistent support path supported by the documented relations. Appendix~\ref{app:support-subgraph-extraction} provides the query-time extraction details.
As shown in Fig.~\ref{fig:osgr-arch}(a), multiple candidates can be extracted simultaneously. Both enter the spatiotemporal grounding phase, while paths failing this logical derivation are filtered directly at the conceptual level, preventing downstream grounding modules from exploring non-valid search spaces. 

\subsection{Relation-Conditioned Grounding}
\label{sec:path-selection}
While support paths ($\mathcal{S}_q$) establish the abstract scientific derivations, they do not determine \textit{which} specific observation items in ($V_{\mathrm{arc}}$) are compatible with the requested location $s_q$ and time $\tau_q$.
In EO, the physical validity is strictly governed by spatial topology (rooted in Region Connection Calculus (RCC) and network routing \citep{randell1992spatial,freeman2007hydrologic}) and temporal topology (governed by Allen's Interval Algebra \citep{allen1983maintaining} for simultaneous or lagged causal feedbacks).
To formalize these theoretical, qualitative relations over discrete observation items, we instantiate spatiotemporal constraints as query-conditioned spatial and temporal compatibility kernels:
\begin{equation}
\label{eq:st-classes}
\begin{aligned}
\Gamma_s
&=\{\texttt{overlap},\texttt{containment},\\
&\quad \texttt{proximity},\texttt{flow-connectivity}\},\\
\Gamma_t
&=\{\texttt{co-temporal},\texttt{lagged-precedence},\\
&\quad \texttt{bi-temporal-change}\}.
\end{aligned}
\end{equation}

For a given query $q$ and active support path $\sigma = (v_0, \dots, v_m)$, the terminal result node $v_m$ specifies the required result variable, while $(q, \sigma)$ determines the active compatibility kernels required for grounding. The grounding process dynamically aggregates the feasible observation items $X_{q,\sigma}$ as:
\begin{equation}
\label{eq:grounded-paths}
\begin{aligned}
&(\gamma_s^{q,\sigma},\gamma_t^{q,\sigma})\in\Gamma_s\times\Gamma_t,\\
X_{q,\sigma}
&=\{x\in V_{\mathrm{arc}}\mid \nu_x=v_m,\\
&\quad F(M_x,q)\bigl[
\gamma_s^{q,\sigma}(\Omega_x,s_q)\\
&\quad+\gamma_t^{q,\sigma}(T_x,\tau_q)
\bigr]\ge\theta_q\},\\
\mathcal{X}^{\mathrm{OSGR}}_q
&=\{X_{q,\sigma}\mid
\sigma\in\mathcal{S}_q,\ X_{q,\sigma}\neq\emptyset\}.
\end{aligned}
\end{equation}
Here, the constraint $\nu_x = v_m$ enforces that each candidate observation item matches the required result variable. As shown in Fig.~\ref{fig:osgr-arch}(b), spatial and temporal compatibility kernels in Eq.~\ref{eq:grounded-paths} are selected dynamically, which provide footprint-normalized local compatibility scores used to form candidate grounded observation-item sets. The final set-level compatibility follows Eq.~\ref{eq:problem}: $C_s(X,s_q)$ and $C_t(X,\tau_q)$ aggregate item-level and relation-level scores, while $F(X,q)$ checks coverage completeness, resolution, provenance, and execution feasibility for the whole selected set. Appendix~\ref{app:relation-kernels} gives the concrete kernel definitions.

Crucially, each non-empty $X_{q,\sigma}$ remains structurally linked to its support path $\sigma$ to ensure downstream answer generation is conditioned on both grounded observation items and the logical reasoning chain.

Under the idealized semantic completeness and scoring equivalence assumptions formally defined in Appendix~\ref{sec:appendix-proof}, the two-stage factorization characterizes when OSGR preserves the feasible family of the original joint search:


\begin{theorem}[Conditional Lossless Factorization]
\label{thm:lossless}
If the support graph is semantically complete and the grounding scores are equivalent to the original compatibility function, then the two-stage factorized search preserves all feasible solutions: $\mathcal{X}^{\mathrm{OSGR}}_q = \mathcal{X}^{\theta}_q$.
\end{theorem}

\begin{proof}[Proof Sketch]
The proof relies on establishing double-inclusion. $\mathcal{X}^{\mathrm{OSGR}}_q \subseteq \mathcal{X}^{\theta}_q$ is guaranteed by scoring equivalence, which ensures any grounded observation-item set meets the original threshold. Conversely, $\mathcal{X}^{\theta}_q \subseteq \mathcal{X}^{\mathrm{OSGR}}_q$ is guaranteed by semantic completeness, which ensures every feasible observation-item set has a corresponding support path. The complete proof is in Appendix~\ref{sec:appendix-proof}.
\end{proof}

\subsection{Complexity and Scalability Analysis}
\label{sec:complexity-scalability}
Rather than enumerating the exponential space of arbitrary subsets in $V_{\mathrm{arc}}$ (scaling as $O(2^{|V_{\mathrm{arc}}|})$), OSGR restricts evaluation strictly to candidate observation-item sets induced by valid support paths. For a path $\sigma=(v_0,\ldots,v_m)$, the localized candidate set is:
\[
\mathcal{C}_{q,\sigma} = \{x\in V_{\mathrm{arc}}\mid \nu_x=v_m\}.
\]
Let $C_q^{\max} = \max_{\sigma\in\mathcal{S}_q}|\mathcal{C}_{q,\sigma}|$. Since the grounding score in Eq.~\ref{eq:grounded-paths} is evaluated only over these candidate sets, the computational complexity of grounding is:
\[
\label{eq:complexity}
O\bigl(|\mathcal{S}_q| \cdot C_q^{\max}\bigr),
\]
which is linear in the candidates evaluated across the extracted support paths rather than enumerating arbitrary observation-item subsets. In practice, metadata indexes and spatiotemporal thresholds prune this candidate space further, yielding reduced growth over the evaluated archive scales.

\definecolor{expblue}{RGB}{52,94,176}
\definecolor{expgreen}{RGB}{52,132,84}
\definecolor{exporange}{RGB}{190,122,38}
\definecolor{expgray}{RGB}{110,110,110}
\definecolor{expred}{RGB}{181,76,59}

\newcommand{\upblue}[1]{\textcolor{expblue}{\scriptsize$\uparrow$#1\%}}
\newcommand{\downblue}[1]{\textcolor{expblue}{\scriptsize$\downarrow$#1\%}}
\newcommand{\upred}[1]{\textcolor{expred}{\scriptsize$\uparrow$#1\%}}
\newcommand{\downred}[1]{\textcolor{expred}{\scriptsize$\downarrow$#1\%}}

\section{Experiments}

\subsection{Experimental Setup}
\paragraph{Datasets.}
We evaluate OSGR on two benchmarks. The primary benchmark is Earth-Bench from Earth-Agent \citep{feng2025earthagent}, which contains 248 expert-curated questions associated with approximately 13.7K EO images and provides expert trajectories used only for evaluation. As a complementary benchmark, we use \textsc{DisasterM3-Reason} based on DisasterM3 \citep{wang2025disasterm3}. It contains 7,160 multiple-choice instances from two disaster-response tasks: 4,982 damaged building counting (BDC) and 2,178 damaged road estimation (DRE).

\paragraph{Evaluation Metrics.}
Following Earth-Bench \citep{feng2025earthagent}, we evaluate the grounded observation-selection pipeline in addition to final answer correctness. We convert each generation into a compact observation-support representation. Evidence Coverage Recall (ECR), Evidence Order Recall (EOR), and Evidence Path Match (EPM) measure whether the predicted path covers, orders, and exactly matches an admissible expert path. Temporal Feasibility Accuracy (TFA) and Spatial Feasibility Accuracy (SFA) evaluate whether the selected EO observations satisfy the query-induced time and space requirements, while Result Grounding Accuracy (RGA) measures whether the answer-supporting result endpoint matches a benchmark-supported endpoint. We also report Efficiency (Eff.) and Answer Accuracy (Acc.). Higher is better for all metrics except Eff., whose ideal value is $1$; values above $1$ indicate redundant execution, whereas values below $1$ can indicate missing necessary steps rather than superior efficiency. Formal definitions are in Appendix~\ref{app:metric-defs}.

\paragraph{Baselines.}
We compare against four baselines. \textbf{Direct LLM} produces a one-shot answer from the raw question without retrieval or graph grounding. \textbf{ReAct Agent} uses the same external execution interface but reasons directly from the question through iterative thought-action-observation loops \citep{yao2023react}. \textbf{GeoGraphRAG} retrieves local subgraphs in the spirit of GeoGraphRAG \citep{liang2025geographrag}. \textbf{Earth-Agent} is the most relevant EO agent baseline on Earth-Bench \citep{feng2025earthagent}.

\begin{table*}[t]
\centering
\small
\resizebox{\textwidth}{!}{
\begin{tabular}{lcccccccc}
\toprule
\textbf{Method} & \textbf{ECR}$\uparrow$ & \textbf{EOR}$\uparrow$ & \textbf{EPM}$\uparrow$ & \textbf{TFA}$\uparrow$ & \textbf{SFA}$\uparrow$ & \textbf{RGA}$\uparrow$ & \textbf{Eff.} $\rightarrow$ 1 & \textbf{Acc.}$\uparrow$ \\
\midrule
\multicolumn{9}{c}{\emph{Closed-Source LLMs}} \\
\midrule
Direct LLM (GPT-4o)              & 56.1 & 27.4 & 22.6  & 25.4 & 22.5 & 20.2 & 0.4 & 21.4 \\
ReAct Agent (GPT-4o)             & 58.7 & 35.2 & 32.4 & 68.1 & 69.3 & 63.9 & 3.8 & 43.6 \\
GeoGraphRAG (GPT-4o)                & 64.4 & 33.5 & 32.1 & 67.9 & 72.1 & 68.7 & 3.6 & 48.8 \\
Earth-Agent (GPT-4o-mini)        & 53.6 & 43.7 & 32.9 & 54.5 & 79.4 & 57.3 & 2.0 & 46.4 \\
Earth-Agent (GPT-4o)             & 61.3 & 44.1 & 33.7 & 57.6 & 78.5 & 69.3 & 2.9 & 49.3 \\
OSGR (GPT-4o-mini)               & 66.4 \upblue{23.9} & \underline{55.5} \upblue{27.0} & 40.3 \upblue{22.5} & 63.2 \upblue{16.0} & \underline{88.0} \upblue{10.8} & 69.4 \upblue{21.1} & 2.1 \upred{5.0} & 58.5 \upblue{26.1} \\
OSGR (GPT-4o)                    & \underline{70.6} \upblue{15.2} & 52.3 \upblue{18.6} & \underline{41.1} \upblue{22.0} & \underline{78.2} \upblue{35.8} & 84.6 \upblue{7.8} & \underline{80.7} \upblue{16.5} & 2.5 \downblue{13.8} & \underline{63.7} \upblue{29.2} \\
OSGR (GPT-5)                     & \textbf{77.3} & \textbf{71.8} & \textbf{60.4} & \textbf{85.7} & \textbf{91.5} & \textbf{83.4} & 2.4 & \textbf{70.9} \\
\midrule
\multicolumn{9}{c}{\emph{Open-Source LLMs}} \\
\midrule
Earth-Agent (Qwen3-MAX)          & 52.5 & 40.4 & 30.5 & 52.4 & 65.9 & 45.3 & 2.6 & 43.3 \\
Earth-Agent (Llama-4)            & 38.8 & 32.5 & 19.8 & 50.9 & 38.4 & 46.5 & 2.7 & 33.9 \\
OSGR (Qwen3-MAX)                 & 59.9 \upblue{14.1} & 47.0 \upblue{16.3} & 35.8 \upblue{17.4} & 64.3 \upblue{22.7} & 74.3 \upblue{12.7} & 50.8 \upblue{12.1} & 2.3 \downblue{11.5} & 56.0 \upblue{29.3} \\
OSGR (Llama-4)                   & 45.7 \upblue{17.8} & 36.4 \upblue{12.0} & 21.4 \upblue{8.1} & 56.2 \upblue{10.4} & 44.1 \upblue{14.8} & 56.8 \upblue{22.2} & 2.1 \downblue{22.2} & 39.2 \upblue{15.6} \\

\bottomrule
\end{tabular}}
\caption{Main comparison on Earth-Bench. Earth-Agent and OSGR are paired under matched backbones. Bold/underline denotes best/second-best for all metrics except Eff. Blue annotations indicate improvements over Earth-Agent with the same backbone, while red annotations indicate degradations.}
\label{tab:earthbench-main}
\end{table*}

\subsection{Main Results on Earth-Bench}
\label{sec:main-results}
Table~\ref{tab:earthbench-main} shows the main comparison results on Earth-Bench, grouped by closed-source and open-source backbone models. Earth-Agent and OSGR are evaluated using matched language models, separating the effect of support-path reasoning and grounding from the basic capabilities of different language models. We observe several clear trends.
\textbf{Limitations of Searching Without Scientific Guidance.}
First, LLMs without external data retrieval perform poorly. {Direct LLM (GPT-4o)} achieves moderate evidence coverage (ECR = 56.1\%) but low grounding accuracy (RGA = 20.2\%) and final accuracy (Acc. = 21.4\%). This indicates that parametric knowledge alone cannot reliably identify the correct observation items. 
While tool-using and graph-based baselines improve accuracy, they lack the coupling of scientific derivation and item-level physical constraints. {ReAct Agent (GPT-4o)} improves temporal and spatial alignments (TFA = 68.1\%, SFA = 69.3\%) via iterative tool use, but over-explores unnecessary observation items, yielding high redundancy ({Eff. = 3.8}). {GeoGraphRAG} reaches 48.8\% answer accuracy, 68.7\% RGA, 32.1\% EPM, and an Eff. of 3.6, whereas OSGR reaches 63.7\%, 80.7\%, 41.1\%, and 2.5, respectively. This matched-backbone gap shows that OSGR is not equivalent to retrieving a local graph neighborhood: the support-path/physical-grounding coupling contributes beyond graph retrieval alone.
\textbf{Performance under Closed-Source Backbones.}
Second, OSGR consistently outperforms Earth-Agent under matched closed-source backbones. Under {GPT-4o-mini}, OSGR improves all path matching metrics (ECR, EOR, and EPM improve by +23.9\%, +27.0\%, and +22.5\% relatively) and raises accuracy by +26.1\% over Earth-Agent. 
Upgrading to {GPT-4o}, OSGR increases accuracy from 49.3\% to {63.7\%} (+14.4 absolute points; +29.2\% relative) and temporal alignment (TFA) from 57.6\% to 78.2\%. Crucially, this gain occurs while the execution-step ratio moves closer to its ideal value of $1$, decreasing from 2.9 to 2.5. This joint improvement demonstrates a better observed accuracy--efficiency trade-off. Finally, {GPT-5} sets the upper performance bound, achieving 60.4\% EPM, 91.5\% SFA, and 70.9\% accuracy.
\textbf{Framework Generalization across Open-Source LLMs.}
Third, the same patterns hold for open-source models, demonstrating model-agnostic robustness within the evaluated EO setting. Under {Qwen3-MAX}, OSGR raises accuracy from 43.3\% to 56.0\% (+29.3\% relative gain), improves RGA to 50.8\%, and lowers Eff. from 2.6 to 2.3. Under the smaller {Llama-4}, OSGR still secures a consistent +15.6\% relative accuracy boost (reaching 39.2\%) and lowers Eff. from 2.7 to 2.1. 
In short, these consistent improvements indicate that OSGR's benefits are not limited to a single language-model family or scale.

\subsection{Search Efficiency and Token Cost Analysis}
\begin{figure}[t]
\centering
\includegraphics[width=\columnwidth]{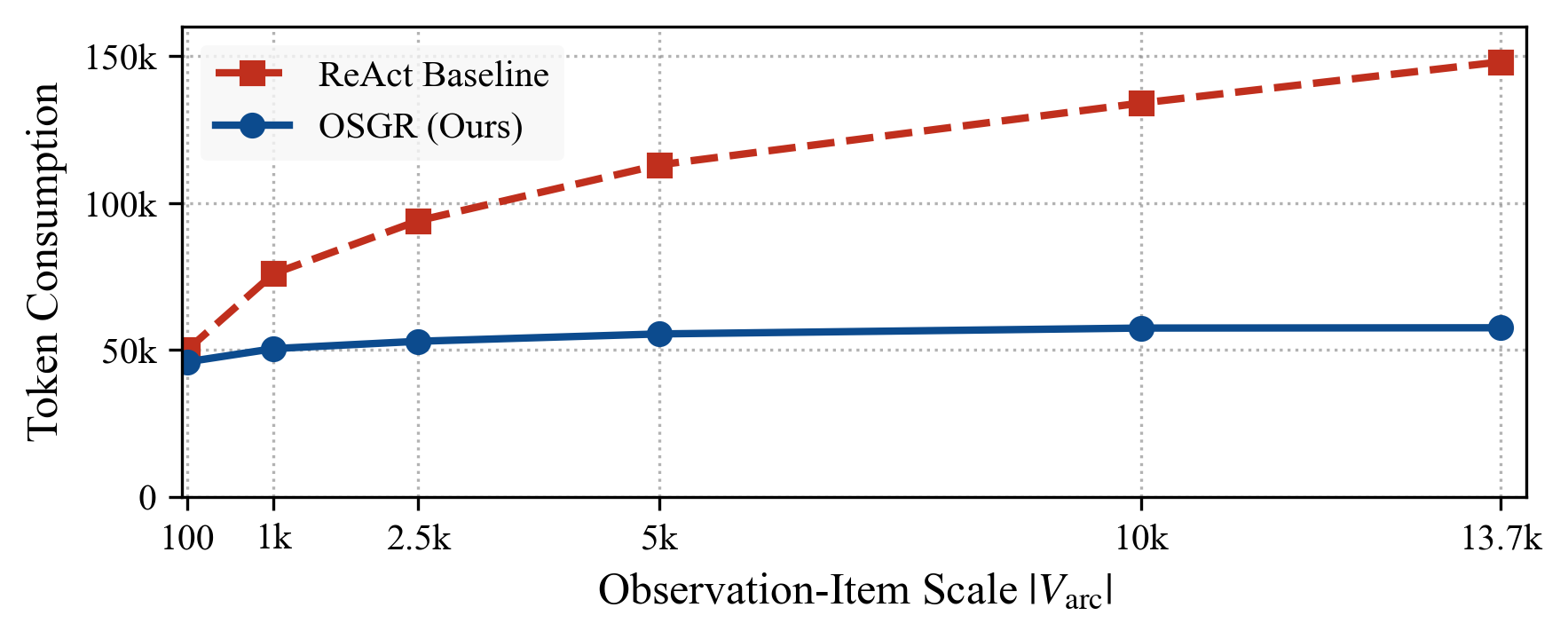}
\caption{Scalability over observation-item scale.}
\label{fig:scale}
\end{figure}

\label{sec:efficiency-analysis}
We evaluate the token efficiency and search-space pruning capability of OSGR across increasing archive scales. Averaged over Earth-Bench, OSGR extracts a highly compact set of only {5.1} abstract support paths per query, which the relation-conditioned grounding stage further reduces to {1.3} grounded observation sets, completely bypassing raw archive enumeration.
Fig.~\ref{fig:scale} plots the observed end-to-end token consumption against the number of candidate files in the physical archive. While the ReAct baseline grows sharply because of lengthy tool-observation histories and iterative error correction, OSGR grows more slowly by exposing compact grounded evidence to the LLM. At the maximum evaluated scale of 13.7K archive items, ReAct consumes roughly 150K tokens, whereas OSGR remains below 60K tokens. This observed gap indicates that delegating low-level spatiotemporal filtering to the database layer reduces the growth of LLM reasoning cost over the evaluated archive scales.

\subsection{Ablation Study}
\begin{table*}[t]
\caption{Ablation study of OSGR core components on Earth-Bench using GPT-4o. Best scores are bolded.}
\label{tab:earthbench-ablation}
\centering
\small
\resizebox{1.6\columnwidth}{!}{
\begin{tabular}{lcccccccc}
\toprule
\textbf{Variant} & \textbf{ECR}$\uparrow$ & \textbf{EOR}$\uparrow$ & \textbf{EPM}$\uparrow$ & \textbf{TFA}$\uparrow$ & \textbf{SFA}$\uparrow$ & \textbf{RGA}$\uparrow$ & \textbf{Eff.} $\rightarrow$ 1 & \textbf{Acc.}$\uparrow$ \\
\midrule
w/o support graph & 46.2 & 33.8 & 29.4 & 67.2 & 80.6 & 52.9 & 4.7 & 43.9 \\
w/o relation-conditioned grounding & \textbf{73.4} & 50.7 & 40.0 & 58.8 & 59.4 & 61.3 & 3.7 & 53.2 \\
\textbf{Full OSGR} & 70.6 & \textbf{52.3} & \textbf{41.1} & \textbf{78.2} & \textbf{84.6} & \textbf{80.7} & \textbf{2.5} & \textbf{63.7} \\
\bottomrule
\end{tabular}
}
\end{table*}

We evaluate OSGR's two claimed core components on Earth-Bench using GPT-4o (Table~\ref{tab:earthbench-ablation}). The stage-level design isolates the division between scientific derivation and physical grounding; neither stage alone explains the full gain.

\paragraph{Impact of the Observation-Support Graph.} 
Removing the support graph prior ({w/o support graph}) disrupts conceptual reasoning, causing path alignment metrics to drop sharply (ECR: 46.2\%, EOR: 33.8\%, EPM: 29.4\%). Without symbolic constraints, the model searches the archive more broadly, inflating execution redundancy (Eff.) to 4.7 and dropping result grounding accuracy (RGA) to 52.9\%. This limits final accuracy to 43.9\%, supporting the role of the graph in filtering invalid concept spaces.

\paragraph{Impact of Relation-Conditioned Grounding.}
Disabling spatiotemporal verification ({w/o relation-conditioned grounding}) yields a deceptively high ECR (73.4\%). Unconstrained by archive-level data feasibility, the symbolic module overfits benchmark text elements. However, these paths fail physical execution: temporal (TFA) and spatial (SFA) feasibility alignments plunge to 58.8\% and 59.4\%, dragging RGA to 61.3\% and accuracy to 53.2\% with an inefficient execution score of 3.7.

\paragraph{Synergistic Division of Labor.}
The contrast establishes a clear division of labor: the support graph eliminates semantically invalid target-to-result derivations at the abstract concept level, while the relation-conditioned grounding determines whether those paths can be physically instantiated under exact spatiotemporal topologies. Coupling both stages allows the full OSGR framework to maximize search efficiency (Eff. = 2.5) and achieve the highest answering accuracy (63.7\%).

\subsection{Error Analysis}
\begin{figure}[t]
\centering
\includegraphics[width=\columnwidth]{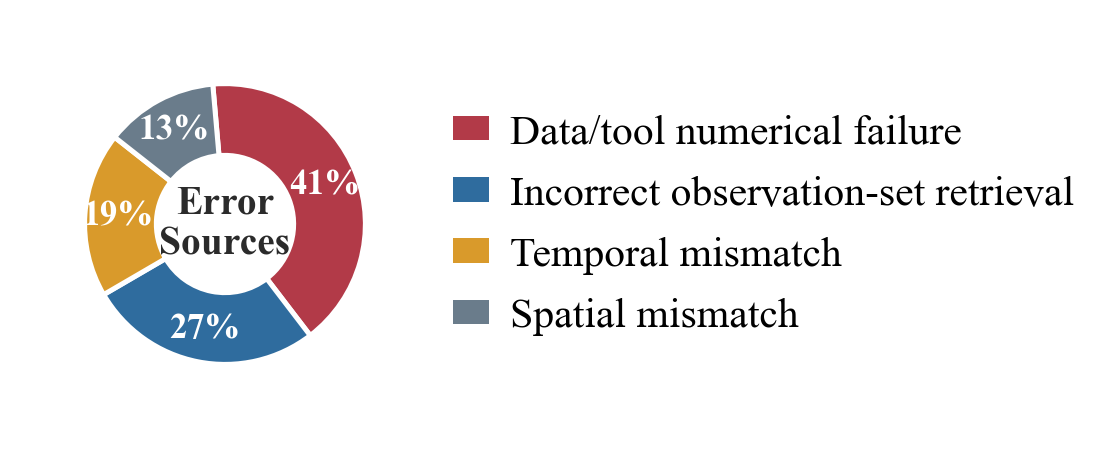}
\caption{Analysis of incorrect OSGR predictions.}
\label{fig:earthbench-error}
\end{figure}

We analyze the failure modes of OSGR (GPT-4o) on Earth-Bench by manually inspecting the incorrect predictions (Fig.~\ref{fig:earthbench-error}). The largest error category comes from data and backend numerical issues (41\%), where selected files contain invalid values like \texttt{NaN} or function parameters do not match the specific data format. Incorrect observation-set retrieval is the second largest source (27\%), which typically occurs when queries are vague or multiple satellite products match the same target concept. Temporal and spatial mismatches account for 19\% and 13\% of the errors, respectively, showing that resolving fine-grained event windows and precise spatial scales remains difficult. This distribution indicates that OSGR successfully minimizes high-level reasoning errors, moving the remaining challenges down to backend execution reliability and data-level precision.

\subsection{Evaluation on DisasterM3-Reason}
\begin{table}[t]
\centering
\small
\resizebox{\columnwidth}{!}{
\begin{tabular}{lccc}
\toprule
Method & \textbf{Avg. Acc.} & BDC & DRE \\
\midrule
\multicolumn{4}{c}{\emph{Closed-Source LLMs}} \\
\midrule
Direct LLM (GPT-4o)           & 28.5 & 28.1 & 29.4 \\
ReAct Agent (GPT-4o)          & 36.0 & 35.6 & 36.8 \\
GeoGraphRAG (GPT-4o)          & 43.1 & 43.2 & 42.8 \\
Earth-Agent (GPT-4o)          & \underline{46.2} & \underline{46.6} & \underline{45.2} \\
OSGR (GPT-4o)                 & \textbf{50.1} \upblue{8.4} & \textbf{49.8} \upblue{6.9} & \textbf{50.7} \upblue{12.2} \\
\midrule
\multicolumn{4}{c}{\emph{Open-Source LLMs}} \\
\midrule
Earth-Agent (Qwen3-MAX)       & \underline{36.3} & \underline{36.4} & \underline{36.1} \\
OSGR (Qwen3-MAX)              & \textbf{39.9} \upblue{9.9} & \textbf{39.1} \upblue{7.4} & \textbf{41.8} \upblue{15.8} \\
\bottomrule
\end{tabular}}
\caption{Results on DisasterM3-Reason. Avg. Acc. is the instance-weighted average over BDC and DRE. Bold/underline denotes best/second-best.}
\label{tab:disasterm3-main}
\end{table}

Table~\ref{tab:disasterm3-main} reports results on {DisasterM3-Reason}, a complementary disaster-response benchmark with a query and task distribution distinct from Earth-Bench.
The results show that OSGR consistently outperforms the baselines across both closed-source and open-source models. Using the {GPT-4o} backbone, OSGR increases the instance-weighted average accuracy from 46.2\% to {50.1\%} (an +8.4\% relative gain). Notably, the performance improvement is larger on the DRE task (+12.2\% relative gain) than on the BDC task (+6.9\% relative gain). This difference occurs because estimating road damage highly depends on comparing pre-disaster and post-disaster observation pairs. OSGR's relation-conditioned grounding uses precise temporal kernels to find these matching pairs, which effectively prevents errors caused by choosing mismatching images.
We observe a similar trend with open-source backbones. Under {Qwen3-MAX}, OSGR raises the average accuracy from 36.3\% to {39.9\%} (+9.9\% relative gain), with a clear advantage on the DRE task (+15.8\% relative gain). This consistent success across different backbones demonstrates that the improvements stem from our structured observation selection and factorized search mechanism rather than the capabilities of a specific language model.

\section{Conclusion}

We presented OSGR, a factorized retrieval-and-derivation framework that decouples symbolic reasoning over an Observation-Support Graph from low-level Relation-Conditioned Grounding over observation items for Earth observation question answering. By resolving scientific validity via abstract support paths and enforcing spatiotemporal compatibility during grounding, OSGR reduces unconstrained search and physical mismatches. Evaluations on two benchmarks show that OSGR consistently improves accuracy and efficiency across multiple LLMs.

\section{Limitations}
OSGR depends on the coverage and correctness of the support graph and archive metadata. Missing support relations, inaccurate footprints, or noisy acquisition times can prevent valid observation-item sets from being recovered. The current empirical evidence is scoped to EO; applying the factorization principle to another scientific archive would require a domain-specific schema and compatibility functions.

\bibliography{custom}
\appendix
\section{Observation-Support Graph Construction}
\label{app:support-graph-construction}

We construct the offline Observation-Support Graph $G_{\mathrm{sup}}$ through an LLM-assisted extraction pipeline grounded in EO scientific documentation. The graph schema follows the Target--Property--Result hierarchy described in the main text and is aligned with the ISO/OGC Observations, Measurements, and Samples framework \citep{ogc2023oms,janowicz2019sosa}. The source corpus contains remote-sensing studies and EO product guides that document how an environmental target is characterized through physical properties and how those properties are measured or estimated from EO results \citep{aldana2022semantic,vandermeer2022semantic}.

\paragraph{Extraction prompt and schema.}
For each source document, we use GPT-4 with the following constrained instruction template. The prompt definitions match the three node types and two edge types in Eqs.~\ref{eq:support-graph}--\ref{eq:support-edges}.

\begin{lstlisting}[
  basicstyle=\ttfamily\scriptsize,
  breaklines=true,
  breakatwhitespace=true,
  breakautoindent=false,
  breakindent=0pt,
  columns=fullflexible,
  keepspaces=true,
  showstringspaces=false
]
Given an Earth-observation scientific document,
extract only Target-Property-Result relations explicitly
supported by it.

Target:
  The environmental phenomenon or scientific object
  studied or inferred (e.g., drought).

Property:
  A physical or derived property used to characterize
  or infer the Target (e.g., TVDI).

Result:
  An observable or computed EO variable, band, index,
  or product used to measure or estimate a Property
  (e.g., MODISA).

Allowed relations:
  has_property:
    Maps a Target node to a Property node.
    Target -> Property

  estimated_from:
    Maps a Property node to another Property node or
    to a Result node.
    Property -> Property
    Property -> Result

Instructions:
  - Extract only relations supported by the document.
  - Preserve standard scientific names and prioritize
    common English acronyms. For example, use 'NDVI'
    instead of 'Normalized Difference Vegetation Index'.
  - Ignore generic data processing steps (e.g., masking,
    resampling, or aggregation).
  - Focus solely on physical or empirical retrieval logic.
  - Return an empty list when no valid relation is stated.
\end{lstlisting}

Each extracted record therefore contains a typed subject, relation, and object. We merge duplicate triples before converting the records into nodes and edges in a Neo4j database. Target, Property, and Result nodes form $V_{\mathrm{tar}}$, $V_{\mathrm{prop}}$, and $V_{\mathrm{res}}$, while \texttt{has\_property} and \texttt{estimated\_from} records form $E_{\mathrm{sup}}$. When several documents support the same semantic relation, the graph stores one edge with multiple provenance records rather than duplicate edges.

Human correction is limited to duplicate-entry checking and merging. Specifically, entries referring to the same concept but separated because of singular/plural forms, abbreviations, or minor naming variants are merged. The graph is consequently an LLM-assisted representation of reusable, documented EO relations rather than question-specific solution paths. Edge-level provenance retains the source document and evidence text or section, which can be retrieved with a support path for downstream traceability.

The support graph is not intended to exhaustively encode all EO science. It organizes stable relations relevant to the target archive, while volatile observation items remain in $G_{\mathrm{arc}}$ and are grounded online. Its one-time construction cost is amortized across queries: new files whose variable types are already represented do not require reconstructing the conceptual graph, and previously unseen relations can be added incrementally.

\section{Support Path Extraction Details}
\label{app:support-subgraph-extraction}
Support path extraction instantiates Eq.~\ref{eq:support-paths} at query time. Starting from the query target $a_q$, extraction follows typed support-graph edges toward archive-available result nodes in $V_{\mathrm{res}}$. A zero-length support path is allowed when $a_q\in V_{\mathrm{res}}$.

The implementation keeps extraction finite using three checks. First, typed traversal admits only schema-consistent transitions implied by \texttt{has\_property} and \texttt{estimated\_from}. Second, cycle avoidance prevents a support path from revisiting the same support node. Third, a result-stop rule stops traversal once a result node is reached, because result nodes are grounded to observation items rather than expanded into more concepts.

\section{Relation-Conditioned Compatibility Kernels}
\label{app:relation-kernels}
This section specifies the spatial and temporal compatibility kernels used to instantiate the abstract scores $C_s$ and $C_t$ in Eq.~\ref{eq:problem}. The main text writes these item-level kernels as $\gamma_s^{q,\sigma}(\Omega_x,s_q)$ and $\gamma_t^{q,\sigma}(T_x,\tau_q)$. In Relation-Conditioned Grounding, the query--path pair $(q,\sigma)$ determines the active kernel type and canonicalizes the raw spatial and temporal cues into the variables required by that kernel. Once these variables are fixed, all compatibility scores are computed deterministically in $[0,1]$.

For spatial grounding, the raw spatial cue $s_q$ is canonicalized into a spatial reference object $R_q$, which can be a point, line, polygon, or multi-polygon. For point and line cues, we use a tolerance buffer
\[
R_q^\rho=\{z\in\mathbb{R}^2\mid d(z,R_q)\le \rho_q\},
\]
where $\rho_q$ is determined by the query scale and product resolution. Polygonal cues can be treated as $R_q^\rho=R_q$ when no buffer is needed. Let $\Omega_x$ denote the footprint of observation item $x$, $|\cdot|$ denote area, $d(\cdot,\cdot)$ denote spatial distance, and $\epsilon$ be a small positive constant. The spatial kernels measure local compatibility and are normalized by the candidate item's footprint or overlap scale, so small but correctly located EO coverage units are not unfairly penalized. To keep equations compact, long kernel labels are abbreviated in superscripts.

The overlap kernel measures shared spatial support when the item and query supports are comparable in scale:
\[
\gamma_s^{\texttt{overlap}}(\Omega_x,s_q)
=
\frac{|\Omega_x\cap R_q^\rho|}
{|\Omega_x\cup R_q^\rho|+\epsilon}.
\]

The containment kernel measures local containment when the item and query supports differ in scale:
\[
\gamma_s^{\texttt{cont}}(\Omega_x,s_q)
=
\frac{|\Omega_x\cap R_q^\rho|}
{\min(|\Omega_x|,|R_q^\rho|)+\epsilon}.
\]

The proximity kernel allows nearby observations when exact overlap is not required:
\[
\gamma_s^{\texttt{prox}}(\Omega_x,s_q)
=
\exp\left(-\frac{d(\Omega_x,R_q)}{\lambda_q^s}\right),
\]
where $\lambda_q^s$ is a query- or product-dependent spatial decay scale.

For flow-connectivity, Euclidean distance is replaced by directed network distance. Let $H$ denote the relevant flow or transport network, and let $d_H^{\delta_q}(\Omega_x,R_q)$ be the shortest path distance from $\Omega_x$ to $R_q$ under the query-required direction $\delta_q$ (e.g., upstream-to-downstream). If no directed path exists, we set $d_H^{\delta_q}(\Omega_x,R_q)=\infty$. Writing $\Delta_H=d_H^{\delta_q}(\Omega_x,R_q)$, the connectivity kernel is
\[
\begin{aligned}
\gamma_s^{\texttt{flow}}(\Omega_x,s_q)
&=
\exp\left(-\frac{\Delta_H}{\lambda_q^H}\right),
\end{aligned}
\]
where $\exp(-\infty)=0$.

Temporal cues are canonicalized analogously. The raw temporal cue $\tau_q$ is mapped to one or more intervals. A time point is treated as a short interval using tolerance $\eta_q$, and an observation acquisition epoch $T_x$ is represented as an interval $[t_x^-,t_x^+]$. For intervals $I$ and $J$, define
\[
d_t(I,J)=
\max(0,\ I^- - J^+,\ J^- - I^+),
\]
and let $o_{IJ}=|I\cap J|$. Define
\[
\begin{aligned}
a_{IJ}
&=\frac{o_{IJ}}{\min(|I|,|J|)+\epsilon},\\
b_{IJ}
&=\exp\left(-\frac{d_t(I,J)}{\lambda}\right).
\end{aligned}
\]
The soft interval-overlap function is
\[
\begin{aligned}
\operatorname{softov}(I,J;\lambda)
&=
\begin{cases}
a_{IJ}, & o_{IJ}>0,\\
b_{IJ}, & o_{IJ}=0.
\end{cases}
\end{aligned}
\]
For compactness, write $\mathrm{so}_q(I,J)=\operatorname{softov}(I,J;\lambda_q^t)$ in the temporal kernels below.

For co-temporal grounding, $\tau_q$ is canonicalized as the requested interval $I_q$:
\[
\gamma_t^{\texttt{co}}(T_x,\tau_q)
=
\mathrm{so}_q(T_x,I_q).
\]

For lagged-precedence grounding, the query specifies a target interval $I_q$ and an admissible lag range $L_q=[\ell_q^-,\ell_q^+]$. The supporting observation should fall in the predecessor window
\[
P_q=[I_q^- - \ell_q^+,\ I_q^+ - \ell_q^-],
\]
so the kernel is
\[
\begin{aligned}
\gamma_t^{\texttt{lag}}(T_x,\tau_q)
&=
\mathrm{so}_q(T_x,P_q)\\
&\quad\cdot
\mathbb{I}\!\left[m(T_x)\le m(I_q)\right],
\end{aligned}
\]
where $m(\cdot)$ denotes the interval midpoint.

For bi-temporal change, $\tau_q$ is canonicalized into a pre-event interval $I_q^{\mathrm{pre}}$ and a post-event interval $I_q^{\mathrm{post}}$. When scoring a single observation item, the support path provides its temporal role $r_x\in\{\mathrm{pre},\mathrm{post}\}$. Let $I_q^{r_x}$ denote the interval selected by that role:
\[
\begin{aligned}
\gamma_t^{\texttt{chg}}(T_x,\tau_q;r_x)
&=
\mathrm{so}_q(T_x,I_q^{r_x}).
\end{aligned}
\]
For a pre/post pair $(x_i,x_j)$, the relation-level score additionally enforces ordering:
\[
\begin{aligned}
\gamma_t^{\texttt{chg}}(T_i,T_j,\tau_q)
&=
\mathrm{so}_q(T_i,I_q^{\mathrm{pre}})\\
&\quad\cdot
\mathrm{so}_q(T_j,I_q^{\mathrm{post}})\\
&\quad\cdot
\mathbb{I}\!\left[m(T_i)<m(T_j)\right].
\end{aligned}
\]

For a candidate observation-item set $X$, the set-level scores are averages over footprint-normalized item-level kernels inherited from the active support path:
\[
\begin{aligned}
C_s(X,s_q)
&=\operatorname{avg}_{x\in X}
\gamma^s_x(\Omega_x,s_q),\\
C_t(X,\tau_q)
&=\operatorname{avg}_{x\in X}
\gamma^t_x(T_x,\tau_q).
\end{aligned}
\]
The spatial kernels measure local compatibility after normalizing by the candidate item's footprint or overlap scale, so small but correctly located coverage units are not unfairly penalized. Coverage completeness and executability at the query scale are handled by $F(X,q)$, which checks whether the selected set $X$ collectively provides sufficient spatial coverage, temporal support, resolution, and required preprocessing. When a support step depends on relations between observation items, such as pre/post change or flow-connected support, the corresponding relation-level score is folded into the same aggregation.

\section{Theoretical Guarantees and Proofs}
\label{sec:appendix-proof}
To state the conditions under which the two-stage factorization is conditionally lossless, we first introduce the axiomatic assumptions underpinning Theorem~\ref{thm:lossless}. Let $X \preceq \sigma$ denote that the variable types of an observation-item set $X$ match the terminal result node of the support path $\sigma$.

\begin{assumption}[Semantic Completeness]
\label{ass:completeness}
Every feasible observation-item set $X \in \mathcal{X}^{\theta}_q$ has a corresponding support path in the offline support graph $G_{\mathrm{sup}}$:
\begin{equation}
\forall X \in \mathcal{X}^{\theta}_q,\, \exists \sigma \in \mathcal{S}_q : X \preceq \sigma.
\end{equation}
\end{assumption}

\begin{assumption}[Scoring Equivalence]
\label{ass:equivalence}
The relation-conditioned spatiotemporal compatibility kernels and metadata verification accurately implement the physical compatibility scoring function:
\begin{equation}
X = X_{q,\sigma} \Longleftrightarrow X \preceq \sigma \land S(X,q) \ge \theta_q.
\end{equation}
\end{assumption}

\begin{remark}[Non-ideal support and metadata]
The theorem is a conditional characterization, not an assertion that a finite practical graph or noisy archive metadata automatically satisfies its assumptions. If semantic completeness fails, valid solutions whose derivations are absent from $G_{\mathrm{sup}}$ may be missed. Under the forward direction of scoring equivalence, however, the grounded sets retained by OSGR remain feasible, so graph incompleteness primarily affects solution recall. Metadata or kernel errors constitute a separate failure source and can affect both false negatives and false positives.
\end{remark}

\subsection{Proof of Theorem~\ref{thm:lossless}}
We present the formal proof of Theorem~\ref{thm:lossless} by demonstrating mutual subset inclusion between the grounded observation-item-set family $\mathcal{X}^{\mathrm{OSGR}}_q$ and the original feasible search space $\mathcal{X}^{\theta}_q$.

\begin{proof}
First, we prove $\mathcal{X}^{\mathrm{OSGR}}_q \subseteq \mathcal{X}^{\theta}_q$. 
For any arbitrary observation-item set $X \in \mathcal{X}^{\mathrm{OSGR}}_q$, there must exist a target-to-result support path $\sigma \in \mathcal{S}_q$ such that $X = X_{q,\sigma}$ and $X \neq \emptyset$. 
By applying the forward implication of Assumption~\ref{ass:equivalence} (Scoring Equivalence), we obtain:
\begin{equation}
X = X_{q,\sigma} \implies X \preceq \sigma \land S(X,q) \ge \theta_q.
\end{equation}
From $S(X,q) \ge \theta_q$, it immediately follows by the definition of the original feasible family in Eq.~\ref{eq:problem} that $X \in \mathcal{X}^{\theta}_q$. Therefore, $\mathcal{X}^{\mathrm{OSGR}}_q \subseteq \mathcal{X}^{\theta}_q$.

Conversely, we prove $\mathcal{X}^{\theta}_q \subseteq \mathcal{X}^{\mathrm{OSGR}}_q$. 
For any arbitrary observation-item set $X \in \mathcal{X}^{\theta}_q$, the definition of the original feasible space dictates that $S(X,q) \ge \theta_q$. 
By Assumption~\ref{ass:completeness} (Semantic Completeness), there must exist a support path $\sigma \in \mathcal{S}_q$ in the pre-constructed offline support graph $G_{\mathrm{sup}}$ such that $X \preceq \sigma$. 
Combining these two components yields:
\begin{equation}
X \preceq \sigma \land S(X,q) \ge \theta_q.
\end{equation}
By applying the reverse implication of Assumption~\ref{ass:equivalence}, this condition directly resolves to:
\begin{equation}
X = X_{q,\sigma}.
\end{equation}
Since $X \in \mathcal{X}^{\theta}_q$, $X$ is non-empty, meaning $X_{q,\sigma} \neq \emptyset$. Thus, by the definition of the grounded observation-item-set family in Eq.~\ref{eq:grounded-paths}, we have $X \in \mathcal{X}^{\mathrm{OSGR}}_q$. This establishes that $\mathcal{X}^{\theta}_q \subseteq \mathcal{X}^{\mathrm{OSGR}}_q$.

Combining both subset containment directions yields $\mathcal{X}^{\mathrm{OSGR}}_q = \mathcal{X}^{\theta}_q$. 
This concludes the proof, establishing losslessness under the two stated assumptions.
\end{proof}

\section{Evaluation Protocol and Metric Definitions}
\label{app:metric-defs}

\paragraph{Evaluation fairness.}
All methods are evaluated on the same benchmark-provided natural-language queries and observation archives. Tool-using methods share the same archive metadata and execution interface, and Earth-Agent and OSGR are paired under matched LLM backbones. The support graph is the method-specific scientific prior being evaluated; it is extracted from EO literature and contains no Earth-Agent code, benchmark answers, gold trajectories, or tool annotations. Thus, OSGR receives no additional answer- or trajectory-level benchmark supervision.

We define the cross-method path metrics with respect to an admissible expert path set
\[
\Sigma_q^\star=\{\sigma_{q,1}^\star,\ldots,\sigma_{q,M_q}^\star\},
\]
where each path $\sigma_q^\star$ is an ordered sequence of support elements drawn from the roles \texttt{property}, \texttt{proxy} (optional), and \texttt{product/result}. Let $\hat \sigma_q$ denote the path extracted from the model prediction and $\mathrm{elem}(\cdot)$ denote its multiset of support elements.

Evidence Coverage Recall (ECR) is
\[
\mathrm{ECR}_q
=
\max_{\sigma_q^\star\in\Sigma_q^\star}
\frac{\big|\mathrm{ms\_int}(\mathrm{elem}(\sigma_q^\star),\mathrm{elem}(\hat\sigma_q))\big|}{|\mathrm{elem}(\sigma_q^\star)|},
\]
where $\mathrm{ms\_int}$ denotes multiset intersection and therefore ignores order but preserves multiplicity. Evidence Order Recall (EOR) is
\[
\mathrm{EOR}_q
=
\max_{\sigma_q^\star\in\Sigma_q^\star}
\frac{\mathrm{LCS}(\sigma_q^\star,\hat \sigma_q)}{|\sigma_q^\star|},
\]
where $\mathrm{LCS}$ is the longest-common-subsequence length and therefore rewards a correct path order. Evidence Path Match (EPM) is
\[
\mathrm{EPM}_q
=
\mathbb{I}\!\left[\hat \sigma_q\in\Sigma_q^\star\right].
\]

Let $\gamma_{t,q}^\star$ and $\gamma_{s,q}^\star$ denote the expert temporal and spatial support-relation labels induced by the query. Temporal Feasibility Accuracy (TFA) and Spatial Feasibility Accuracy (SFA) are
\[
\begin{aligned}
\mathrm{TFA}_q&=\mathbb{I}\!\left[\hat\gamma_{t,q}=\gamma_{t,q}^\star\right],\\
\mathrm{SFA}_q&=\mathbb{I}\!\left[\hat\gamma_{s,q}=\gamma_{s,q}^\star\right].
\end{aligned}
\]

Let $\hat d_q$ denote the answer-supporting result endpoint selected by the model and let $\mathcal{D}_{\mathrm{sup}}(q)$ denote the set of benchmark-supported endpoints for the query. Result Grounding Accuracy (RGA) is
\[
\mathrm{RGA}_q
=
\mathbb{I}\!\left[\hat d_q\in\mathcal{D}_{\mathrm{sup}}(q)\right].
\]

Efficiency (Eff.) is defined as
\[
\mathrm{Eff}_q = \frac{n_q}{m_q},
\]
where $n_q$ and $m_q$ are the numbers of predicted and expert execution steps under the shared backend. A value of $1$ is ideal, values above $1$ indicate redundant tool use, and values below $1$ may indicate under-execution rather than superior efficiency. Answer Accuracy (Acc.) is verified with a deterministic answer checker:
\[
\mathrm{Acc}_q = \mathbb{I}\!\left[V_{\mathrm{ans}}(\hat a_q,a_q)=1\right].
\]
Dataset-level Earth-Bench scores are reported as the mean of each per-query metric under the auto-planning setting.

Task-level answer metrics on \textsc{DisasterM3-Reason} use multiple-choice accuracy for all retained tasks,
\[
\mathrm{Acc}^{\mathrm{dm3}}_q
=
\mathbb{I}\!\left[\hat y_q = y_q\right].
\]
This follows the multiple-choice evaluation protocol used for damaged building counting and damaged road estimation in the original DisasterM3 benchmark. The reported task averages on \textsc{DisasterM3-Reason} are means over the corresponding per-query accuracies.

\end{document}